\documentclass{article}

\usepackage{iclr2025_conference,times}
\iclrfinalcopy

\usepackage{amsmath,amsfonts,bm}

\def\eqref#1{equation~\ref{#1}}

\def\1{\bm{1}}

\DeclareMathAlphabet{\mathsfit}{\encodingdefault}{\sfdefault}{m}{sl}
\SetMathAlphabet{\mathsfit}{bold}{\encodingdefault}{\sfdefault}{bx}{n}

\usepackage{amsmath}
\usepackage{amssymb}
\usepackage{graphicx}
\usepackage{booktabs}
\usepackage[table]{xcolor}
\usepackage{multirow}
\usepackage{makecell}
\usepackage{adjustbox}
\usepackage{wrapfig}
\usepackage{algorithm}
\usepackage{algpseudocode}
\usepackage{hyperref}
\usepackage{url}

\usepackage{caption}
\title{TRACE: Transition-Aware Residual Control for Multi-Objective Materials Discovery}

\author{
Kang Zhou\textsuperscript{1}\thanks{Equal contribution.},
Yujia Tong\textsuperscript{1}\footnotemark[1]\thanks{Corresponding author.},
Yong Tao\textsuperscript{1},
Jingling Yuan\textsuperscript{1}\footnotemark[2] \\
\textsuperscript{1} Wuhan University of Technology \\
{\tt\small \{tyjjjj, yjl\}@whut.edu.cn}\\
}

\begin{document}

\maketitle

\begin{abstract}
Multi-objective materials discovery with LLM agents is often limited not only by how many candidates can be proposed, but by how effectively each costly property evaluation informs the next search step. Existing agents mainly store evaluated candidates and their scores, so they know which materials succeeded but not which executable edits caused useful property changes. This makes local refinement difficult when objectives compete and an edit that improves one property may damage another. We propose TRACE, a transition-aware residual control framework that treats evaluated edits as the basic unit of feedback. TRACE records each local refinement as a parent-edit-child transition with observed property deltas, aggregates transition evidence to estimate reusable edit effects, and ranks future edits by their predicted ability to reduce the current candidate’s remaining constraint violations while avoiding damage to already satisfied objectives. In a controlled same-backbone comparison, TRACE improves over LLEMA, the state-of-the-art LLM-agent baseline, raising macro-average hit rate from 18.13\% to 25.96\%.

\end{abstract}

\section{Introduction}
\label{sec:introduction}

Multi-objective materials discovery is essential for identifying viable materials that simultaneously satisfy multiple, often competing, property requirements in increasingly demanding scientific and engineering applications ~\citep{butler2018machine,oganov2019structure,gopakumar2018multi}. 
Recent advances in LLM-based agents have enabled domain knowledge and property evaluation tools to be integrated into iterative search processes, allowing agents to propose, evaluate, and refine candidate materials more  efficiently ~\citep{white2023future,m2024augmenting}.

Despite this progress, existing LLM-based materials discovery agents primarily utilize feedback at the candidate level, retaining evaluated structures, property values, or textual summaries of success and failure ~\citep{jia2024llmatdesign,takahara2025accelerated,abhyankar2026llema}. Such feedback tells the agent whether a candidate performs well, but does not explicitly reveal which executable modification was associated with the observed property changes, making it difficult to determine how the candidate should be improved next. This limitation is particularly consequential in multi-objective discovery, where a modification that improves one property may degrade another~\citep{jablonka2021bias,daulton2020differentiable}, and the usefulness of that modification depends on which constraints of the current candidate remain unsatisfied~\citep{zeng2026expert}.

\begin{figure}[t]
    \centering
    \includegraphics[width=\linewidth]{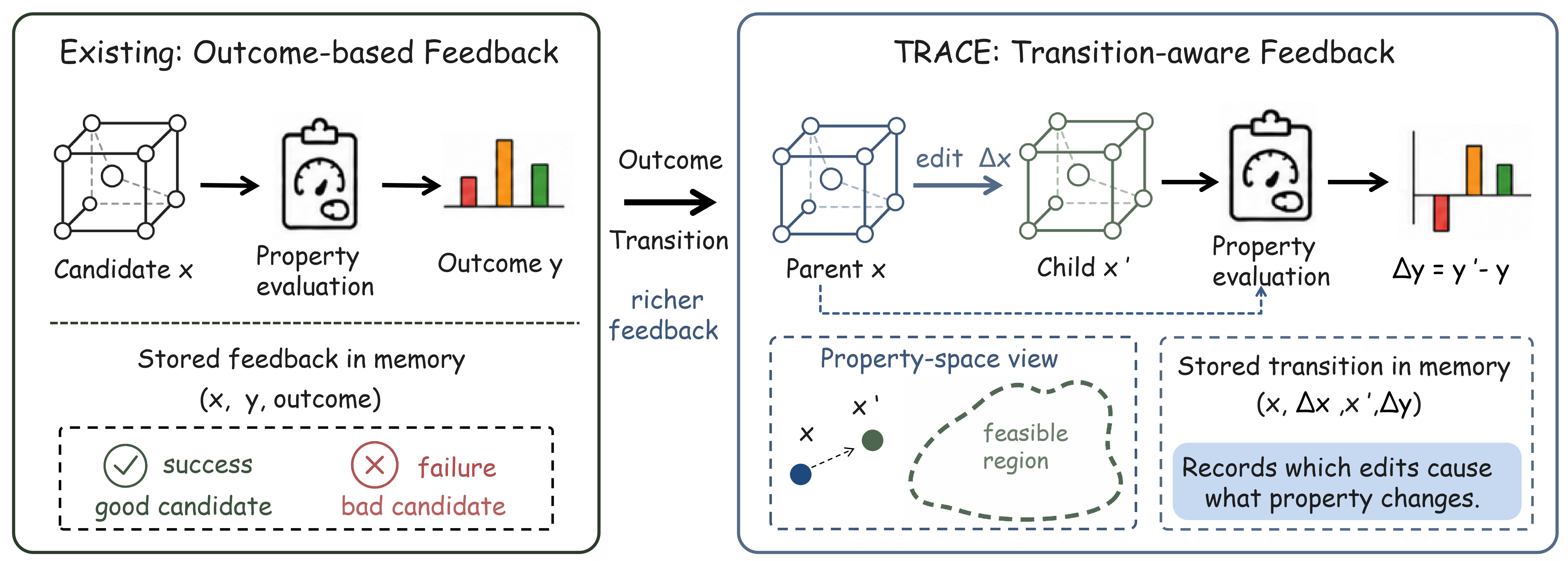}
    \setlength{\abovecaptionskip}{-5pt}
    \caption{Motivation. Candidate-level feedback records which candidates perform well, whereas transition-aware feedback captures how executable edits change material properties and uses this evidence to guide subsequent search.}
    \label{fig:motivation}
\end{figure}

Figure~\ref{fig:motivation} makes this distinction explicit. Treating each evaluation as a transition provides reusable evidence about what a particular
modification may achieve. Previously observed edit--property relationships can then help identify modifications aligned with the current candidate's remaining unsatisfied objectives. The key question is how to convert this evidence into concrete decisions about what to edit next. Motivated by this insight, we propose TRACE, a transition-aware residual control framework that operationalizes such transition feedback for multi-objective materials discovery. TRACE records each evaluated parent–edit–child transition together with the resulting property changes, enabling the agent to estimate how executable edits are likely to affect task-relevant properties from accumulated search experience. It then selects edits according to the current candidate’s remaining constraint violations, favoring modifications that address unsatisfied objectives while avoiding degradation of properties that already meet their targets. By combining this feedback-driven local editing with global LLM exploration, TRACE exploits previously observed edit effects while retaining the ability to explore new regions of the material space.

We evaluate TRACE on 14 multi-objective materials discovery tasks from LLEMABench against a broad range of generative and agent-based baselines, with controlled comparisons to LLEMA under approximately matched property-evaluation counts. TRACE improves over LLEMA under matched backbones and approximately matched property-evaluation counts, increasing the Qwen macro-average hit rate from 18.13\% to 25.96\%. These results demonstrate that explicitly reusing transition-level feedback can translate into more effective multi-objective materials search.
Our main contributions are summarized as follows:

\begin{itemize}

    \setlength{\labelsep}{5pt}      
    \item \textbf{Transition-level feedback formulation.}
     We formulate material-search feedback as reusable parent--edit--child transition experience, explicitly associating executable edits with observed property changes rather than relying solely on candidate-level outcomes.

    \item \textbf{Transition-aware residual control.}
     We develop TRACE, which integrates transition memory, lightweight edit-effect estimation, and residual-aware edit selection. Historical edit effects are prioritized according to current constraint residuals, while global LLM exploration preserves search beyond the local executable edit space.

    \item \textbf{Comprehensive empirical evaluation.}
    We evaluate TRACE across 14 multi-objective LLEMABench tasks under controlled property-evaluation budgets, with component ablations, exploration-allocation studies, and transition-level analyses to validate both discovery performance and the effectiveness of transition feedback
    
\end{itemize}

\section{Problem Formulation}
\label{sec:problem_formulation}

\subsection{Oracle-Limited Multi-Objective Materials Discovery}

Let $x\in\mathcal{X}$ denote a candidate material and let an oracle
$\mathcal{O}$ return its task-relevant property vector
\begin{equation}
    \mathbf{y}(x)=\mathcal{O}(x)
    =[y_1(x),\ldots,y_m(x)].
    \label{eq:oracle}
\end{equation}
The oracle may be a simulator, learned surrogate, or experimental measurement~\citep{low2024evolution,abhyankar2026llema}.
Only query access to its outputs is required. A task specifies a set of property constraints
$\mathcal{C}=\{c_j\}_{j=1}^{m}$. Each  constraint $c_j$ defines a
feasible set $\mathcal{F}_j\subseteq\mathbb{R}$, such as $[t_j,\infty)$ for a
lower bound, $(-\infty,t_j]$ for an upper bound, or $[l_j,u_j]$ for an
interval. We define its normalized violation as
\begin{equation}
    d_j(x)=\frac{\operatorname{dist}\!\left(y_j(x),\mathcal{F}_j\right)}{s_j},
    \label{eq:constraint_residual}
\end{equation}
where $\operatorname{dist}(y,\mathcal{F})=\inf_{z\in\mathcal{F}}|y-z|$ and
$s_j>0$ places heterogeneous properties on comparable scales. The total
constraint residual is
\begin{equation}
    D(x;\mathcal{C})=\sum_{j=1}^{m}d_j(x).
    \label{eq:total_residual}
\end{equation}
The notation $D(\mathbf{y};\mathcal{C})$ applies the same calculation directly
to a property vector. Task-specific validity checks handle non-numerical
composition and safety requirements. We set $h(x;\mathcal{C})=1$ when $x$
passes these checks and all property constraints, and $h(x;\mathcal{C})=0$
otherwise. Given limited oracle evaluations, the objective is to discover candidates with $h(x;\mathcal{C})=1$.

\subsection{Executable Edits and Transition Feedback}

The search agent may generate a new candidate globally or apply an executable edit $e\in\mathcal{E}(x)$ to an evaluated parent $x$. The edit operator deterministically constructs a child
\begin{equation}
    x'=\mathcal{G}(x,e),
    \label{eq:executable_edit}
\end{equation}
which is evaluated by the same oracle. Edits must be enumerable, executable,
and compatible with the material representation.  A completed and evaluated edit produces the transition
\begin{equation}
    \tau=(x,e,x',\mathbf{y}(x),\mathbf{y}(x'),\Delta\mathbf{y}),
    \qquad
    \Delta\mathbf{y}=\mathbf{y}(x')-\mathbf{y}(x).
    \label{eq:transition}
\end{equation}
Its realized residual gain is
\begin{equation}
    g_D(\tau)=D(x;\mathcal{C})-D(x';\mathcal{C}),
    \label{eq:residual_gain}
\end{equation}
where $g_D>0$ indicates movement toward the feasible region. Candidate-level
feedback retains $(x',\mathbf{y}(x'))$, whereas transition feedback additionally links the observed property difference to the executed edit. An edit contributes transition evidence only when its child is successfully constructed and evaluated.

\section{TRACE: Transition-Aware Residual Control}
\label{sec:method}

\begin{figure*}[t]
    \centering
    \includegraphics[width=\linewidth]{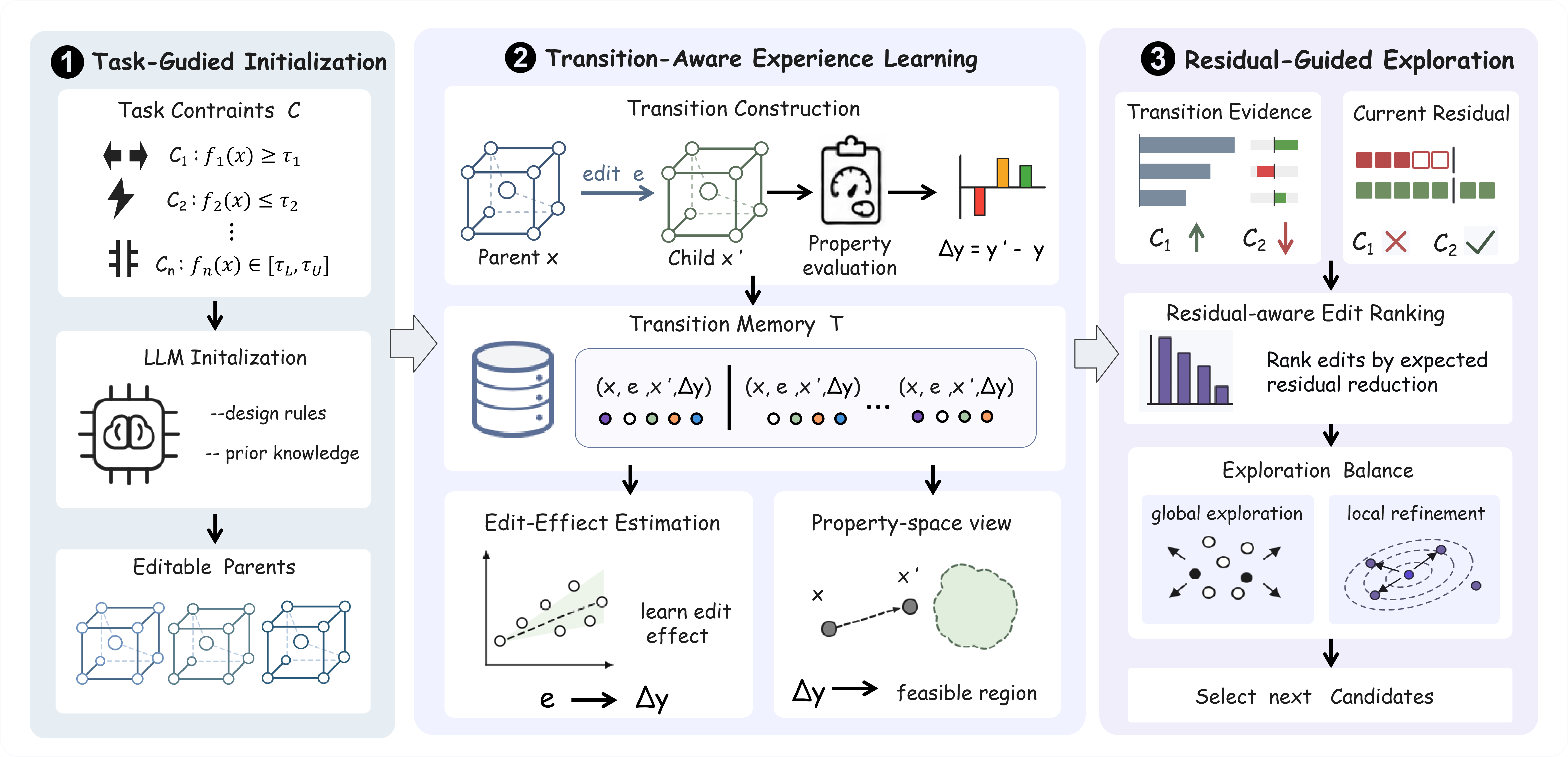}
    \caption{Overview of TRACE. Global LLM proposals support broad exploration,
    while local Delta edits refine evaluated parents according to their current
    constraint residuals. Evaluated parent--edit--child transitions update
    transition memory, whose accumulated property changes guide subsequent edit
    selection.}
    \label{fig:trace_framework}
\end{figure*}

TRACE augments an LLM-guided evolutionary search loop with transition-level feedback and residual-aware local control. As shown in
Figure~\ref{fig:trace_framework}, TRACE begins from an editable parent pool initialized by the LLM under the task constraints. Transition memory $\mathcal{T}$ then accumulates parent--edit--child transitions from local edits on evaluated parents and supplies the evidence for reusable
edit-effect estimation. During search, these edit effects are matched with the current constraint residuals to rank local Delta edits, while the global LLM route maintains broader exploration through task-guided proposals

\subsection{Transition Memory}

Transition memory is built online from executed local edits. After iteration
$t$, it contains
\begin{equation}
    \mathcal{T}_t
    =\{\tau_i\}_{i=1}^{n_t}
    =\{\tau_i:\mathbf{y}(x'_i)\text{ is observed}\},
    \label{eq:transition_memory}
\end{equation}
where each $\tau_i$ follows Eq.(\ref{eq:transition}). Only completed edits
enter $\mathcal{T}_t$, because $\Delta\mathbf{y}_i$ requires an observed child
property vector. Unsuccessful executions remain in an auxiliary attempt log
but provide no property-difference supervision. We condition all transition
sets on the current task and omit the task index.

Each completed transition is annotated with the parent and child constraint
states. Let $q(x)=\sum_{j=1}^{m}\mathbb{I}[d_j(x)=0]$ denote the number of
satisfied numerical constraints and $v(x)=\mathbb{I}[D(x;\mathcal{C})=0]$
denote numerical feasibility. Unlike $h(x;\mathcal{C})$ in
Section~\ref{sec:problem_formulation}, $v(x)$ does not include non-numerical
validity checks. TRACE records the residual gain $g_D(\tau_i)$ and assigns the
transition-level utility
\begin{equation}
    u_i=\lambda_v b_i
        +\lambda_q\bigl(q(x'_i)-q(x_i)\bigr)
        +\lambda_D g_D(\tau_i),
    \label{eq:transition_utility}
\end{equation}
where $b_i$ rewards entering or preserving numerical feasibility and penalizes
leaving it. With nonnegative weights, $u_i$ combines boundary behavior, the
change in satisfied constraints, and continuous residual progress. It therefore
separates an edit's observed property change from its value for the current
constraint state.

\subsection{Estimating Reusable Edit Effects}

TRACE uses a lightweight empirical estimator rather than a full parent-conditioned dynamics model. Let $k(e)$ denote an exact edit signature
and $o(e)$ its operator class. For a transition subset $A$, let
$\boldsymbol{\mu}_{\Delta}(A)$ and $\mu_u(A)$ denote its mean property delta
and transition utility. We construct $\mathcal{T}_{k(e)}$ from transitions with
the same exact signature and $\mathcal{T}_{o(e)}$ from transitions with the
same operator. The predicted edit effect uses a hierarchical backoff:
\begin{equation}
\widehat{\Delta\mathbf{y}}_t(e)=
\begin{cases}
\boldsymbol{\mu}_{\Delta}(\mathcal{T}_{k(e)}),
    & |\mathcal{T}_{k(e)}|\ge n_{\min},\\
\boldsymbol{\mu}_{\Delta}(\mathcal{T}_{o(e)}),
    & |\mathcal{T}_{k(e)}|<n_{\min},\ |\mathcal{T}_{o(e)}|>0,\\
\boldsymbol{\mu}_{\Delta}(\mathcal{T}_t), & \text{otherwise}.
\end{cases}
\label{eq:hierarchical_edit_effect}
\end{equation}
The historical utility $\widehat{u}_t(e)$ is estimated with the same exact-edit, operator, and global backoff. This construction pools evidence when an exact substitution is rare, while retaining edit-specific statistics once sufficient observations are available. When $\mathcal{T}_t$ is empty, the estimates are left unavailable and the local route uses stochastic selection.

For an evaluated parent $x$, an edit induces the projected properties
\begin{equation}
    \widehat{\mathbf{y}}(x,e)
    =\mathbf{y}(x)+\widehat{\Delta\mathbf{y}}_t(e),
    \label{eq:projected_properties}
\end{equation}
and the predicted residual progress
\begin{equation}
    \widehat{g}_D(x,e)
    =D(x;\mathcal{C})
     -D(\widehat{\mathbf{y}}(x,e);\mathcal{C}).
    \label{eq:predicted_progress}
\end{equation}
The estimates are refreshed as transitions accumulate. Before enough evidence
is available, TRACE selects executable edits stochastically. In all cases, the
projection is used only for ranking: the oracle result, rather than the
projection, is written back to transition memory.

\subsection{State- and Residual-Aware Edit Selection}

\paragraph{Parent-state assessment.}
TRACE separates where to edit from which edit to execute. It characterizes a
parent by $\mathbf{s}(x)=(v(x),q(x),D(x;\mathcal{C}))$ and groups evaluated
candidates into numerically feasible, near-feasible, partial, and distant
states. Editable numerically feasible and near-feasible parents receive greater
priority, while other states retain exploratory access and repeatedly
unproductive parents are downweighted.

\paragraph{Residual-aware edit ranking.}
For a selected parent, TRACE enumerates unseen executable edits
$\mathcal{E}(x)$. For numerically feasible, partial, and distant parents, it
evaluates each edit with the general quality score
\begin{equation}
    S_{\mathrm{qual}}(x,e)
    =\widehat{u}_t(e)
     +\lambda_g\widehat{g}_D(x,e)
     +\lambda_n N(x,e),
    \label{eq:quality_edit_score}
\end{equation}
Here $\widehat{u}_t(e)$ summarizes historical transition utility,
$\widehat{g}_D(x,e)$ measures parent-specific residual progress, and $N(x,e)$
rewards an unexplored child composition or pattern.

Near the feasibility boundary, total residual reduction alone can hide an
important failure: an edit may repair the last violated objective while
breaking one that was already satisfied. We therefore split the constraints into
$\mathcal{M}(x)=\{j:d_j(x)>0\}$ and $\mathcal{S}(x)=\{j:d_j(x)=0\}$ and define
\begin{equation}
\begin{aligned}
    G_{\mathrm{miss}}(x,e)
    &=\sum_{j\in\mathcal{M}(x)}
      \left[d_j(x)-d_j\!\left(\widehat{\mathbf{y}}(x,e)\right)\right],
      \\
    B_{\mathrm{sat}}(x,e)
    &=\sum_{j\in\mathcal{S}(x)}
      d_j\!\left(\widehat{\mathbf{y}}(x,e)\right).
\end{aligned}
\label{eq:residual_components}
\end{equation}
$G_{\mathrm{miss}}$ measures progress on violated constraints, whereas
$B_{\mathrm{sat}}$ measures damage to constraints already satisfied. TRACE
therefore scores near-feasible edits as
\begin{equation}
\begin{split}
    S_{\mathrm{near}}(x,e)={}&
    \lambda_m G_{\mathrm{miss}}(x,e)
    -\lambda_b B_{\mathrm{sat}}(x,e)\\
    &+\lambda_f\mathbb{I}
      [D(\widehat{\mathbf{y}}(x,e);\mathcal{C})=0]
    +\lambda_u\widehat{u}_t(e)+\lambda_n N(x,e).
    \label{eq:near_edit_score}
\end{split}
\end{equation}
The feasibility indicator rewards crossing the remaining constraint boundary;
the other terms retain historical evidence and novelty. We write
$S_x=S_{\mathrm{near}}$ for near-feasible parents and
$S_x=S_{\mathrm{qual}}$ otherwise.

Operationally, historical utility first forms a stochastic shortlist
$\mathcal{A}_K(x)\subseteq\mathcal{E}(x)$, with probabilities proportional to
$\exp(\widehat{u}_t(e)/T)$. TRACE reranks it using the score associated with
$\mathbf{s}(x)$ and retains a small exploration probability outside the leading
edits.

\subsection{Coordinating Global and Local Exploration}

Local transition evidence is necessarily concentrated around previously
evaluated material families and cannot reach beyond the executable edit space.
Global LLM proposals move across compositions and families but provide no
explicit edit-level credit. TRACE therefore divides scheduled candidate-attempt
capacity between the two routes. For an iteration with $b_t$ scheduled
candidate-attempt slots, let
\begin{equation}
    \bigl(b_t^{\mathrm{G}},b_t^{\mathrm{L}}\bigr)
    \propto (r_{\mathrm{G}},r_{\mathrm{L}}),
    \qquad
    b_t^{\mathrm{G}}+b_t^{\mathrm{L}}=b_t,
    \label{eq:global_local_budget}
\end{equation}
where $r_{\mathrm{G}}:r_{\mathrm{L}}$ is the planned global--local ratio. The
split is a search-control parameter, not a requirement that both routes produce
the same number of successfully evaluated candidates.

The information flow is bidirectional. Global proposals that pass evaluation
enter candidate memory and may later become editable parents. Conversely,
transition summaries provide the LLM with soft guidance about useful and
harmful edits. If the local route cannot complete its assigned candidate
attempts, global generation receives the unused slots. We study this allocation
trade-off in Section~\ref{sec:experiments}.

\subsection{Overall Algorithm}

The overall TRACE procedure coordinates global LLM exploration with transition-guided local refinement under a scheduled-attempt budget. Both routes share the same construction and oracle-evaluation pipeline, while
evaluated local edits are returned to transition memory as parent--edit--child evidence for subsequent edit-effect estimation. The complete procedure is summarized in Algorithm~\ref{alg:trace}, with additional operational details provided in Appendix~\ref{app:trace_operations}.

\section{Experiments}
\label{sec:experiments}

We organize the evaluation around five questions: (Q1) Does TRACE improve multi-objective discovery under approximately matched property-evaluation counts? (Q2) Does it convert a fixed attempt budget into valid discoveries more efficiently? (Q3) Which components account for the improvement? (Q4) How does the allocation between global and local exploration affect search performance? (Q5) Does
transition history contain predictive and actionable edit-effect information?

\subsection{Experimental Setup}

\paragraph{Benchmark and tasks.}
We evaluate TRACE on the 14 multi-objective materials discovery tasks in LLEMABench~\citep{abhyankar2026llema}. The suite covers electronic, dielectric, energy, mechanical, acoustic, and structural applications, with heterogeneous property and composition constraints.
We retain the benchmark's task definitions, constraint thresholds, candidate-construction pipeline, and property evaluators. Candidate properties are retrieved from reference data when a matching material is available and otherwise estimated with the task-specific surrogate oracle used by LLEMABench.

\paragraph{Baselines and backbones.}
We include the benchmark-reported crystal generators CDVAE~\citep{xie2022crystal},
G-SchNet~\citep{gebauer2019symmetry}, DiffCSP~\citep{jiao2023crystal}, and
MatterGen~\citep{zeni2025generative}, as well as the LLM-based baselines End2end, 
LLMatDesign~\citep{jia2024llmatdesign}, and
LLEMA~\citep{abhyankar2026llema}. The closest controlled comparison pairs LLEMA and TRACE using either Qwen2.5-14B-Instruct or
Mistral-Small-3.2-24B-Instruct-2506 as a shared 8-bit-quantized backbone, with five runs for each task--backbone combination. We use quantized backbones\cite{tong2025robust,tong2025data} to reduce inference cost and enable controlled multi-run evaluation under practical compute budgets. We vary the number of search iterations to target approximately 150 candidates entering property evaluation per method and run. The non-LLM results retain the original LLEMABench protocol of 1,500
generated samples. Since generated samples and candidates entering property evaluation correspond to different stages of the candidate pipeline, the non-LLM results are included as broad reference points rather than budget-matched comparisons. Additional configuration details, iteration counts, and baseline budgets are provided in Appendix~\ref{app:evaluation_llemabench}.

\paragraph{Metrics and accounting.}
The main benchmark and ablations report hit rate (H.R.) and stability rate (Stab.) over candidates that enter property evaluation; Stab. further requires an energy above hull below $0.1$ eV/atom. Generation rate (G.R.) and the long-horizon analysis instead account for all scheduled attempts, including construction and pre-evaluation failures as misses. Complete metric definitions and accounting rules are provided in Appendix~\ref{app:evaluation_metrics}.

\begin{table*}[t]
\vspace{-0.4em}
\centering
\scriptsize
\setlength{\tabcolsep}{2pt}
\renewcommand{\arraystretch}{0.95}
\setlength{\abovecaptionskip}{2pt}
\setlength{\belowcaptionskip}{3pt}
\caption{Comparison of TRACE with baselines on 14 LLEMABench materials discovery tasks. We report hit rate (H.R.) and stability rate (Stab.), where higher values indicate better performance. Best values are bold and gray-highlighted; second-best values are underlined.}
\label{tab:main_results}
\begin{adjustbox}{width=\textwidth}
\begin{tabular}{l*{14}{c}}
\toprule
\textbf{Method}
& \multicolumn{2}{c}{\makecell{Wide-Bandgap\\Semiconductors}}
& \multicolumn{2}{c}{\makecell{SAW/BAW\\Acoustic Substrates}}
& \multicolumn{2}{c}{\makecell{High-k\\Dielectrics}}
& \multicolumn{2}{c}{\makecell{Solid-State\\Electrolytes}}
& \multicolumn{2}{c}{\makecell{Piezo Energy\\Harvesters}}
& \multicolumn{2}{c}{\makecell{Transparent\\Conductors}}
& \multicolumn{2}{c}{\makecell{Insulating\\Dielectrics}} \\
\cmidrule(lr){2-3}\cmidrule(lr){4-5}\cmidrule(lr){6-7}
\cmidrule(lr){8-9}\cmidrule(lr){10-11}\cmidrule(lr){12-13}
\cmidrule(lr){14-15}
& H.R. & Stab. & H.R. & Stab. & H.R. & Stab. & H.R. & Stab.
& H.R. & Stab. & H.R. & Stab. & H.R. & Stab. \\
\midrule
CDVAE
& 0.04 & 0.04 & 0.29 & 0.00 & 0.82 & 0.00 & 0.04 & 0.04
& 42.19 & 0.00 & 0.00 & 0.00 & 1.06 & 0.12 \\
G-SchNet
& 0.00 & 0.00 & 0.42 & 0.00 & 0.00 & 0.00 & 0.00 & 0.00
& 0.01 & 0.00 & 2.49 & 0.00 & 0.01 & 0.00 \\
DiffCSP
& 0.00 & 0.00 & 0.36 & 0.00 & 0.75 & 0.00 & 0.00 & 0.00
& 41.21 & 0.00 & 0.01 & 0.00 & 1.13 & 0.04 \\
MatterGen
& 6.56 & 4.15 & 26.27 & 0.00 & 0.64 & 0.00 & 5.33 & 3.11
& 21.64 & 0.00 & 9.38 & 0.00 & 0.91 & 0.10 \\
End2end
& 0.95 & 0.79 & 10.32 & 0.65 & 0.00 & 0.00 & 0.49 & 0.30
& 10.34 & 0.28 & 0.00 & 0.00 & 0.00 & 0.00 \\
LLMatDesign
& 4.19 & 1.13 & \cellcolor{gray!20}\bfseries 47.59 & 0.13 & 1.35 & 0.32 & 2.51 & 2.44
& 32.16 & 1.38 & 0.04 & 0.04 & 0.21 & 0.08 \\
\midrule
LLEMA (Qwen)
& 25.23 & 18.52 & 30.19 & \underline{8.08} & 7.70 & 4.51 & 28.97 & 23.40
& 63.93 & 6.10 & 22.73 & 4.21 & 10.48 & 4.41 \\
TRACE (Qwen)
& \underline{44.78} & \underline{30.94} & 36.39 & 7.44 & 5.73 & 1.27
& \underline{50.19} & \underline{38.51}
& 61.04 & \underline{15.45} & \cellcolor{gray!20}\bfseries 39.62 & 8.90 & \cellcolor{gray!20}\bfseries 18.07 & 4.11 \\
\midrule
LLEMA (Mistral)
& \cellcolor{gray!20}\bfseries 51.98 & \cellcolor{gray!20}\bfseries 43.67 & 27.54 & 6.65 & \underline{11.09} & \cellcolor{gray!20}\bfseries 5.28 & 31.96 & 27.35
& \cellcolor{gray!20}\bfseries 69.98 & 7.77 & 21.62 & \underline{10.01} & 15.32 & \underline{8.75} \\
TRACE (Mistral)
& 36.59 & 29.50 & \underline{43.21} & \cellcolor{gray!20}\bfseries 14.70 & \cellcolor{gray!20}\bfseries 12.38 & \underline{4.85} & \cellcolor{gray!20}\bfseries 60.06 & \cellcolor{gray!20}\bfseries 48.15
& \underline{66.23} & \cellcolor{gray!20}\bfseries 15.53 & \underline{27.65} & \cellcolor{gray!20}\bfseries 11.26 & \underline{16.62} & \cellcolor{gray!20}\bfseries 9.59 \\
\midrule
\textbf{Method}
& \multicolumn{2}{c}{\makecell{Photovoltaic\\Absorbers}}
& \multicolumn{2}{c}{\makecell{Hard Coating\\Materials}}
& \multicolumn{2}{c}{\makecell{Hard, Stiff\\Ceramics}}
& \multicolumn{2}{c}{\makecell{Aerospace\\Materials}}
& \multicolumn{2}{c}{\makecell{Acousto-optic\\Hybrids}}
& \multicolumn{2}{c}{\makecell{Low-Density\\Structures}}
& \multicolumn{2}{c}{\makecell{Perovskite\\Oxides}} \\
\cmidrule(lr){2-3}\cmidrule(lr){4-5}\cmidrule(lr){6-7}
\cmidrule(lr){8-9}\cmidrule(lr){10-11}\cmidrule(lr){12-13}
\cmidrule(lr){14-15}
& H.R. & Stab. & H.R. & Stab. & H.R. & Stab. & H.R. & Stab.
& H.R. & Stab. & H.R. & Stab. & H.R. & Stab. \\
\midrule
CDVAE
& 1.07 & 0.00 & 0.00 & 0.00 & 15.25 & 0.11 & 1.18 & 0.00
& \cellcolor{gray!20}\bfseries 21.85 & 0.00 & 0.00 & 0.00 & 0.00 & 0.00 \\
G-SchNet
& 0.00 & 0.00 & 0.00 & 0.00 & 0.20 & 0.20 & 0.06 & 0.00
& 0.01 & 0.00 & 0.17 & 0.00 & 0.04 & 0.00 \\
DiffCSP
& 1.11 & 0.00 & 0.00 & 0.00 & 14.75 & 0.00 & 0.09 & 0.00
& \underline{21.53} & 0.01 & 0.00 & 0.00 & 0.04 & 0.00 \\
MatterGen
& 1.88 & 0.00 & 2.12 & 0.00 & 8.23 & 0.00 & 7.34 & 0.00
& 11.24 & 0.00 & 0.27 & 0.00 & 0.93 & 0.00 \\
End2end
& \underline{24.59} & \underline{10.72} & 0.00 & 0.00 & 14.27 & 5.13 & 0.00 & 0.00
& 8.57 & 0.64 & 1.99 & 0.40 & 0.00 & 0.00 \\
LLMatDesign
& 3.92 & 0.00 & 0.00 & 0.00 & 19.00 & 0.41 & 0.00 & 0.00
& 15.45 & 0.55 & 0.07 & 0.00 & 1.10 & 0.81 \\
\midrule
LLEMA (Qwen)
& 23.48 & 4.09 & 0.13 & 0.00 & 6.08 & 2.90 & 6.28 & 1.96
& 16.12 & 2.12 & 2.35 & 0.68 & 10.19 & 1.56 \\
TRACE (Qwen)
& \cellcolor{gray!20}\bfseries 31.32 & 5.08 & 0.26 & 0.13 & 27.08 & 4.06 & 8.33 & \cellcolor{gray!20}\bfseries 2.16
& 19.82 & 3.28 & 3.08 & 0.64 & \cellcolor{gray!20}\bfseries 17.77 & \underline{3.56} \\
\midrule
LLEMA (Mistral)
& 21.78 & \cellcolor{gray!20}\bfseries 13.52 & \cellcolor{gray!20}\bfseries 9.91 & \cellcolor{gray!20}\bfseries 2.92 & \underline{52.32} & \cellcolor{gray!20}\bfseries 19.87 & \cellcolor{gray!20}\bfseries 9.05 & \underline{2.11}
& 11.67 & \underline{4.36} & \underline{16.63} & \underline{2.14} & 4.60 & \cellcolor{gray!20}\bfseries 3.68 \\
TRACE (Mistral)
& 22.96 & 8.96 & \underline{7.28} & \underline{2.17} & \cellcolor{gray!20}\bfseries 57.42 & \underline{11.59} & 5.92 & 1.13
& 14.28 & \cellcolor{gray!20}\bfseries 8.32 & \cellcolor{gray!20}\bfseries 19.07 & \cellcolor{gray!20}\bfseries 3.30 & \underline{12.70} & 2.45 \\
\bottomrule
\end{tabular}
\end{adjustbox}
\vspace{-0.4em}
\end{table*}

\subsection{Main Results on Multi-Objective Materials Discovery}

Table~\ref{tab:main_results} reports task-level H.R. and Stab. Under controlled LLEMA--TRACE comparisons with the task, backbone, and oracle fixed and approximately matched numbers of candidates entering property evaluation, TRACE improves H.R. over LLEMA on 12 of 14 tasks with Qwen and 10 of 14 tasks with Mistral. The macro-average H.R. increases from 18.13\% to 25.96\% with Qwen and from 25.39\% to 28.74\% with Mistral, while the corresponding Stab. averages increase from 5.90\% to 8.97\% and from 11.29\% to 12.25\%. The largest cross-backbone gains occur on Solid-State Electrolytes, where TRACE improves H.R. by 21.22 and 28.10 percentage points and Stab. by 15.11 and 20.80 points with Qwen and Mistral, respectively; Transparent Conductors also improves on both metrics under both backbones. Performance nevertheless remains task dependent, with decreases on several task--backbone settings and mixed H.R./Stab. changes on Piezo Energy Harvesters. Overall, TRACE improves discovery performance on the majority of tasks with both backbones, demonstrating broad effectiveness across the benchmark.

\subsection{Search-Budget Efficiency and Convergence}

\begin{figure*}[t]
    \centering
    \includegraphics[width=\textwidth]{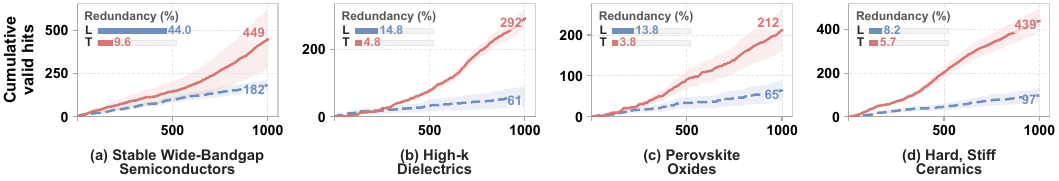}
    \caption{Search-budget efficiency across four representative tasks. Curves and shaded regions show the mean and variability of cumulative valid discoveries over five runs; insets report exact-formula redundancy.}
    \label{fig:sample_efficiency}
\end{figure*}

Final H.R. alone does not capture how quickly valid candidates are discovered. We therefore compare chronological LLEMA and TRACE search trajectories over 1,000 scheduled attempts on four representative tasks, counting construction and pre-evaluation failures as misses. Figure~\ref{fig:sample_efficiency} shows that TRACE accumulates more valid
candidates at the 250, 500, 750, and 1,000-attempt checkpoints on all four tasks. At the final checkpoint, valid discoveries increase from 182 to 449 for Wide-Bandgap Semiconductors, 61 to 292 for High-k Dielectrics, 97 to 439 for Hard, Stiff Ceramics, and 65 to 212 for Toxic-Free Perovskite Oxides, corresponding to $2.47$--$4.79\times$ more discoveries under the same scheduled-attempt budget. Alongside these gains, TRACE yields lower formula redundancy on all four tasks, including reductions from 43.96\% to 9.58\% on Wide-Bandgap Semiconductors and from 14.75\% to 4.79\% on High-k Dielectrics. It also completes more property evaluations within the fixed attempt budget. Together, higher evaluation completion and stronger constraint satisfaction allow TRACE to convert the same scheduled-attempt budget into substantially more valid discoveries while preserving formula diversity.

\subsection{Component Ablation}
\label{sec:ablation}

\begin{wraptable}[8]{r}{0.49\textwidth}
\centering
\scriptsize
\setlength{\tabcolsep}{2.0pt}
\renewcommand{\arraystretch}{1.03}
\setlength{\abovecaptionskip}{0pt}
\setlength{\belowcaptionskip}{0pt}
\vspace*{-12 pt}
\caption{Component ablation of TRACE on Solid-State Electrolytes. Values are mean $\pm$ sample standard deviation over five runs.}
\label{tab:ablation}
\resizebox{\linewidth}{!}{%
\begin{tabular}{lccc}
\toprule
Variant & H.R. $\uparrow$ & Stab. $\uparrow$ & G.R. $\uparrow$ \\
\midrule
Candidate-level & $37.29\pm15.10$ & $24.90\pm8.55$ & $85.13\pm4.87$ \\
Random edits & $32.94\pm13.28$ & $24.35\pm9.10$ & $95.75\pm1.62$ \\
History-guided & $40.15\pm10.30$ & $28.28\pm9.54$ & $\mathbf{97.88\pm1.69}$ \\
TRACE & $\mathbf{50.19\pm14.45}$ & $\mathbf{38.51\pm8.69}$ & $96.63\pm1.44$ \\
\bottomrule
\end{tabular}%
}
\end{wraptable}

We ablate TRACE on Solid-State Electrolytes with Qwen2.5-14B-Instruct under a fixed scheduled budget. 
The ablation in Table~\ref{tab:ablation} follows a progressive sequence from candidate-level feedback to random editing, history-guided exploration, and full TRACE, thereby isolating the effects of executable editing, transition history, and residual-aware selection. Relative to candidate-level feedback, random editing increases G.R. by 10.62 percentage points, while H.R. decreases by 4.35 points and Stab. remains nearly unchanged, indicating that executable editing mainly improves generation completion. Adding transition-history guidance improves H.R. and Stab. over random editing by 7.21 and 3.93 points, respectively. Full TRACE further improves H.R. by 10.04 points and Stab. by 10.23 points while maintaining a 96.63\% G.R. 
These results show that executable editing improves search completion, while transition history and residual-aware selection provide the main gains in discovery quality.

\subsection{Balancing Global and Local Exploration}

\begin{wrapfigure}[16]{r}{0.49\textwidth}
    \vspace{-0.7\baselineskip}
    \vspace*{-10 pt}
    \centering
    \includegraphics[width=\linewidth]
    {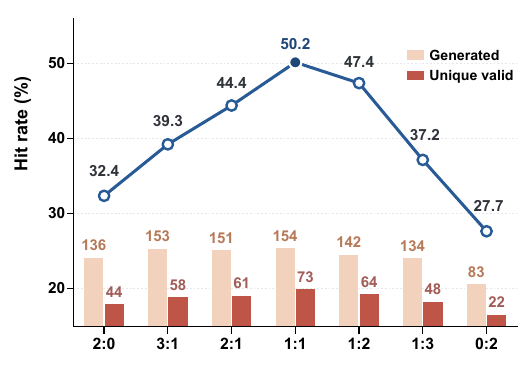}
    \vspace{-25pt}
    \caption{Sensitivity to the LLM:Delta budget ratio on Solid-State Electrolytes. Lines report mean H.R.; bars report generated structures and unique valid formulas.} 
    \label{fig:edit_frequency}
\end{wrapfigure}

Using the same Solid-State Electrolytes task and 8-bit-quantized Qwen2.5-14B-Instruct backbone as in the component ablation, we vary the planned global--local budget ratio while holding all other components and the total scheduled budget fixed. The allocation study in Figure~\ref{fig:edit_frequency} evaluates LLM:Delta ratios of $2\!:\!0$, $3\!:\!1$, $2\!:\!1$, $1\!:\!1$, $1\!:\!2$, $1\!:\!3$, and $0\!:\!2$, with five runs per setting. H.R. increases from 32.4\% with LLM-only exploration to 50.19\% at the balanced $1\!:\!1$ allocation, then declines as local editing dominates, reaching 27.7\% with Delta-only exploration. The balanced setting also yields 154 structures and 73 unique valid formulas, compared with 83 and 22 under Delta-only exploration. The two endpoints expose the limitations of either route alone: LLM-only exploration lacks transition-guided local refinement, whereas Delta-only exploration is restricted to the local edit space. Overall, coordinating global and local exploration yields stronger search performance than relying on either route alone, with the balanced $1\!:\!1$ setting performing best among the tested allocations.

\subsection{Analysis of Learned Edit Effects}

\begin{table*}[t]
\centering
\scriptsize
\renewcommand{\arraystretch}{1.10}
\caption{Transition-level analysis of 882 post-warm-up transitions across four tasks and five runs per task, reporting $\Delta$-direction accuracy, offline-replay selection advantage, and pooled formula uniqueness.}
\label{tab:edit_effect}
\vspace{-6 pt}

\begin{tabular*}{\textwidth}{@{\extracolsep{\fill}}lcccccc@{}}
\toprule
Task & Runs & $N_{\mathrm{edit}}$
& \makecell{$\Delta$-Dir.\\Acc. (\%)}
& \makecell{Selection\\Adv. (\%)}
& \makecell{Valid\\Yield (\%)}
& \makecell{Valid Formula\\Uniq. (\%)} \\
\midrule
Wide-Bandgap Semiconductors & 5 & 218 & 76.49 & 59.66 & 62.75 & 81.75 \\
High-k Dielectrics          & 5 & 216 & 61.12 & 64.16 & 9.62  & 100.00 \\
Solid-State Electrolytes    & 5 & 253 & 69.09 & 54.31 & 67.06 & 89.47 \\
Perovskite Oxides           & 5 & 195 & 70.99 & 57.51 & 38.63 & 85.71 \\
\bottomrule
\end{tabular*}
\end{table*}

The preceding experiments evaluate end-to-end search performance; here we directly examine whether transition history contains predictive and actionable edit-effect information. We analyze 882 post-warm-up transitions from 20 complete runs across four representative tasks.
$\Delta$-direction accuracy measures agreement between predicted and observed property-change directions, while selection advantage compares TRACE's top-ranked edit with a random alternative through offline replay, with 50\% as the random reference. Direction accuracy exceeds 60\% on all four tasks, ranging from 61.12\% to 76.49\%, and selection advantage ranges from 54.31\% to 64.16\%, consistently above the random reference.

The selected edits also produce valid and diverse outcomes. Valid yield ranges from 9.62\% on High-k Dielectrics to 67.06\% on Solid-State Electrolytes and exceeds 60\% on Wide-Bandgap Semiconductors and Solid-State Electrolytes. Pooled valid-formula uniqueness remains between 81.75\% and 100\%, corresponding to limited cross-run overlap despite independent search trajectories. Together, these results support the predictive and actionable value of edit-effect information derived from transition history, while successful edits remain diverse across runs. Additional repetition statistics are provided in Appendix~\ref{app:formula_repetition}.

\section{Related Work}
\label{sec:related_work}

\subsection{LLM Agents for Scientific and Materials Discovery}
LLM agents couple language-based reasoning with scientific tools and external feedback to support iterative discovery~\citep{kang2024chatmof,ghafarollahi2025sciagents}. Coscientist combines planning, search, code execution, and laboratory automation, whereas CHEMREASONER uses quantum-chemical feedback to steer catalyst hypothesis generation~\citep{boiko2023autonomous,sprueill2024chemreasoner}. Similar agentic frameworks have been applied to materials discovery: LLMatDesign refines materials using predicted properties, while MatAgent augments inorganic crystal generation with memory and property feedback~\citep{jia2024llmatdesign,takahara2025accelerated}. Most closely related, LLEMA combines LLM proposals, chemistry-aware evolutionary operators, oracle evaluation, and candidate memory for multi-objective search~\citep{abhyankar2026llema}. These systems typically preserve search experience as reflections, scores, or candidate-level memories, leaving the effects of individual modifications implicit. TRACE instead records evaluated parent--edit--child transitions and reuses the resulting transition evidence to guide subsequent search.

\subsection{Materials Inverse Design and Evolutionary Search}
Learning-based inverse design aims to generate materials satisfying target specifications~\citep{sanchez2018inverse}. CDVAE learns latent representations of periodic crystals, DiffCSP jointly diffuses lattice and atomic variables, and MatterGen supports conditional generation under chemical and property constraints~\citep{xie2022crystal,jiao2023crystal,zeni2025generative}. Sequential search methods use prior evaluations to decide what to explore next~\citep{lookman2019active,montoya2020autonomous,sun2024large}. Multi-objective active learning prioritizes Pareto-efficient regions under limited simulation budgets, while MOLLEO introduces chemistry-aware LLM operators for evolutionary molecular search~\citep{cheng2017coevolutionary,jablonka2021bias,wang2025efficient}. These approaches focus mainly on generating or selecting promising candidates, but provide limited information about the expected effects of specific local modifications. TRACE complements them by using observed edit--property effects to guide local refinement.

\subsection{Transition Modeling and Feedback-Guided Search}
In oracle-limited optimization, efficiency depends on how evaluation outcomes are reused~\citep{chandra2017co,tripp2020sample}. PMO highlights the importance of fixed oracle budgets, while Augmented Memory improves sample efficiency by replaying scored molecules across policy updates~\citep{gao2022sample,guo2024augmented}. 
Other molecular optimization methods model local transformations more directly: graph-to-graph translation learns property-improving transformations from paired examples, Modof performs fragment-level molecular modification, and MARS selects fragment edits for multi-objective search~\citep{jin2018learning,chen2021deep,xie2021mars}. Model-based reinforcement learning provides a related perspective by learning action-conditioned dynamics for planning or policy optimization~\citep{lim2012adaptation,chua2018deep,janner2019trust,cao2023bayesian}. 
Yet the observed effects of executed material edits remain implicit in subsequent search. TRACE makes this evidence explicit through a lightweight transition formulation for one-step refinement,
without full dynamics modeling or multi-step rollouts.

\section{Conclusion}
\label{sec:conclusion}

We introduced \textbf{TRACE}, a transition-aware residual control framework for oracle-limited, multi-objective materials discovery. TRACE records parent--edit--child transitions and reuses their observed property changes to guide local edits according to the current candidate's remaining constraint residuals, while retaining global LLM proposals for broader exploration.
Across 14 LLEMABench tasks and two LLM backbones, TRACE improves hit rate over LLEMA on most tasks under approximately matched property-evaluation counts. Component and transition-level analyses further show that transition history provides actionable guidance beyond executable editing alone. Overall, our results suggest that evaluated transitions serve as a useful feedback unit for guiding oracle-limited materials search.

\bibliography{references}
\bibliographystyle{iclr2025_conference}


\appendix

\section{Operational Details of TRACE}
\label{app:trace_operations}

This appendix specifies the runtime layer beneath the abstract TRACE
description, from constructing the editable parent pool and admissible local
edits to activating transition-guided selection after cold start.
Algorithm~\ref{alg:trace} assembles these choices into the complete search
procedure. The estimator and residual-aware scores follow the definitions in
the main paper.

\subsection{Search State and Memory}
\label{app:search_state_memory}

TRACE constructs a candidate pool before assigning island seeds. It first
generates five formulas and builds their structures; each successfully
constructed candidate is then evaluated with the task oracle. After receiving these
outcomes, TRACE generates a second batch of five formulas using the measured
properties as task feedback and the previously proposed formulas as an
exclusion set. The evaluated pool is stratified as valid, boundary, or poor
according to task-constraint status and normalized residual. Up to two valid
candidates are used as feasible anchors, and the remaining island slots are
filled with exploration seeds chosen for compositional diversity, editability,
and proximity to the feasible region. This role composition is adaptive:
boundary or poor candidates can seed islands when valid anchors are scarce,
whereas valid candidates can also serve as exploration seeds when the pool is
uniformly feasible.

After initialization, local search relies on two additional stores. The
fresh-valid frontier temporarily prioritizes newly discovered, editable valid
candidates from either the global or local route. Transition memory instead
stores evaluated parent--edit--child records and their observed property
changes. The frontier identifies where local search can continue, whereas
transition memory records how executed edits changed material properties.
Frontier entries expire after limited use, while candidate and transition
stores are maintained under fixed capacity settings.

\subsection{Executable Local Editing}
\label{app:executable_local_editing}

A candidate can serve as a local parent only when its lattice, atomic species,
and fractional coordinates are available. The current implementation enumerates
same-group elemental substitutions for the distinct species in that parent. It
filters out unsafe target elements, previously executed parent--edit pairs, and
edits whose child would duplicate an evaluated formula before action selection.

Executing a substitution replaces every site occupied by the selected source
element while preserving the parent lattice, fractional coordinates, and
number of sites. The child formula is recomputed, inherited property values are
discarded, and the resulting structure is passed through the same construction
and oracle-evaluation pipeline used by global proposals. Only a
successfully evaluated child contributes an observed transition. The present
operator therefore implements structure-preserving elemental substitution; it
does not cover vacancy creation, lattice reconstruction, or multi-step reaction
paths.

\subsection{Online Decision Process}
\label{app:online_decision_process}

Transition memory is empty at the start of a run. During this cold-start phase,
TRACE samples from safe, unseen substitutions that pass the executable-edit
filters, rather than relying on estimated edit effects. Each evaluated child adds its
observed property change to transition memory. TRACE enables learned guidance only
after the warm-up period has passed and the minimum transition count has been
reached. TRACE then periodically refreshes the effect estimates and applies the
shortlist and residual-aware ranking defined in the main paper.

An edit is executed and evaluated before the next local decision, so predicted
effects guide ranking but do not replace the oracle outcome stored in memory.
If no eligible parent or admissible edit is available, the local route produces
no candidate and its unused scheduled capacity is reassigned to global
generation.
Appendix~\ref{app:experimental_configuration} reports the numerical settings
for activation, estimator refresh, frontier lifetime, and parent cooldown.

\subsection{Complete TRACE Procedure}
\label{app:complete_trace_procedure}

Algorithm~\ref{alg:trace} summarizes the full TRACE search loop. Starting from
an evaluated initial pool, TRACE initializes candidate memory $\mathcal{M}$,
transition memory $\mathcal{T}$, and the task-valid discovery set
$\mathcal{H}$. At each iteration, scheduled candidate attempts are allocated
between global LLM proposals and transition-guided local edits. Edit-effect
estimates are updated from the accumulated transition memory, while candidates
from both routes share the same construction and oracle-evaluation pipeline.
Successfully evaluated candidates are added to $\mathcal{M}$, and those
satisfying $h(x;\mathcal{C})=1$ are further included in $\mathcal{H}$.
Evaluated local edits append their parent--edit--child transitions to
$\mathcal{T}$, whereas global proposals contribute transition evidence only
after being selected as parents and subsequently edited. When $\mathcal{T}$ is
sparse, local edit selection remains stochastic and becomes increasingly
evidence-guided as transition experience accumulates.

\begin{algorithm}[H]
\caption{TRACE search under budget $B_{\mathrm{sched}}$}
\label{alg:trace}
\begin{algorithmic}[1]

\Require Task $\mathcal{C}$, oracle $\mathcal{O}$, LLM proposer $L$,
edit operator $\mathcal{G}$, budget $B_{\mathrm{sched}}$
\Ensure Valid discoveries $\mathcal{H}$, transition memory $\mathcal{T}$

\State Initialize $\mathcal{M}$, $\mathcal{T}\leftarrow\varnothing$, and $\mathcal{H}$

\While{$B_{\mathrm{sched}}$ is not exhausted}

    \State Compute $(b_t^{\mathrm{G}},b_t^{\mathrm{L}})$
    by Eq.~\ref{eq:global_local_budget}
    \State Update $\widehat{\Delta\mathbf{y}}_t$ and $\widehat{u}_t$
    from $\mathcal{T}_t$

    \Statex $\triangleright$ \textit{Transition-guided local search}
    \State Select parent $x$ by $\mathbf{s}(x)$
    \State Enumerate $\mathcal{E}(x)$ and form $\mathcal{A}_K(x)$
    \State Rank $\mathcal{A}_K(x)$ by $S_x$
    according to Eqs.~\ref{eq:quality_edit_score}--\ref{eq:near_edit_score}
    \State Select $\mathcal{A}_t^{\mathrm{L}}$,
    $|\mathcal{A}_t^{\mathrm{L}}|\le b_t^{\mathrm{L}}$
    \State $\mathcal{X}_t^{\mathrm{L}}
    \leftarrow\{\mathcal{G}(x,e):e\in\mathcal{A}_t^{\mathrm{L}}\}$

    \Statex $\triangleright$ \textit{Global exploration}
    \State $\mathcal{X}_t^{\mathrm{G}}
\leftarrow L(\mathcal{M},\operatorname{summary}(\mathcal{T}), b_t^{\mathrm{G}})$

   \Statex $\triangleright$ \textit{Evaluation and memory update}
    \State Evaluate $\mathcal{X}_t^{\mathrm{L}}\cup
    \mathcal{X}_t^{\mathrm{G}}$ with $\mathcal{O}$
    \State Update $\mathcal{M}$ and $\mathcal{H}$
    \State Append evaluated local transitions to $\mathcal{T}$

\EndWhile

\State \Return $\mathcal{H},\mathcal{T}$

\end{algorithmic}
\end{algorithm}

\section{Benchmark Tasks and Evaluation Metrics}
\label{app:evaluation_protocol}

This section specifies the benchmark scope, comparison protocol, and metric
accounting used in the experiments.

\subsection{Benchmark Scope and Matched Comparisons}
\label{app:evaluation_llemabench}

We use the 14 application-oriented, multi-objective materials-discovery tasks
in LLEMABench~\citep{abhyankar2026llema} as a standardized evaluation suite.
These tasks provide target applications, property constraints, and associated
domain rules for determining whether a candidate satisfies the requested
design objective. TRACE is evaluated as an independently specified search
framework, with its search state, transition memory, edit-effect estimation,
and residual-aware decision policy defined in the main paper.

For the closest controlled comparison, LLEMA and TRACE are independently
instantiated on the same tasks with the same language-model backbone and
evaluation criteria. The shared backbones are Qwen2.5-14B-Instruct and
Mistral-Small-3.2-24B-Instruct-2506, both run in 8-bit quantized form to reduce
inference cost and support controlled multi-run evaluation~\citep{tong2025robust,tong2025data}.
Each matched task--backbone setting is averaged over five runs, with each run
targeting approximately 150 candidates that enter property evaluation. Because
pre-evaluation completion is stochastic and differs between methods, LLEMA uses
85 search iterations and TRACE uses 75 to obtain approximately matched
evaluated-candidate counts. Results for non-LLM generators retain the original
LLEMABench budget of 1,500 samples and serve as benchmark reference points
rather than budget-matched comparisons.

\subsection{Evaluation Metrics and Attempt Accounting}
\label{app:evaluation_metrics}

Let $S$ denote scheduled candidate attempts and let
$\mathcal{X}_{\mathrm{eval}}=\{x_i\}_{i=1}^{N}$ contain the $N\leq S$
candidates that enter property evaluation. The evaluation-conditional hit rate
is
\begin{equation}
    \mathrm{HR}_{\mathrm{eval}}
    =\frac{1}{N}\sum_{i=1}^{N}h(x_i;\mathcal{C}),
    \label{eq:evaluated_hit_rate}
\end{equation}
where $h(x_i;\mathcal{C})=1$ if $x_i$ satisfies all task constraints. Stability
uses the same denominator and additionally requires energy above hull below
$0.1$ eV/atom:
\begin{equation}
    \mathrm{Stab}_{\mathrm{eval}}
    =\frac{1}{N}\sum_{i=1}^{N}
      h(x_i;\mathcal{C})
      \mathbb{I}\!\left[E_{\mathrm{hull}}(x_i)<0.1\ \mathrm{eV/atom}\right].
    \label{eq:evaluated_stability_rate}
\end{equation}
Generation rate instead uses scheduled attempts as its denominator:
\begin{equation}
    \mathrm{GR}=\frac{N}{S}.
    \label{eq:generation_rate}
\end{equation}

These metrics separate two stages of the discovery pipeline.
$\mathrm{HR}_{\mathrm{eval}}$ and $\mathrm{Stab}_{\mathrm{eval}}$ characterize
candidate quality conditional on reaching property evaluation: the former
requires all task constraints, whereas the latter additionally applies the
stability criterion. In contrast, $\mathrm{GR}$ measures generation
completion, or the fraction of scheduled attempts that produce fully evaluable
candidates, rather than candidate validity. Reporting these quantities separately avoids
interpreting high conditional validity as high end-to-end efficiency when many
attempts fail before evaluation.

The long-horizon experiment retains every scheduled attempt and measures
end-to-end discovery yield as
\begin{equation}
    \mathrm{Yield}_{\mathrm{sched}}
    =\frac{1}{S}\sum_{s=1}^{S}\widetilde{h}_s,
    \qquad
    \widetilde{h}_s=
    \begin{cases}
        h(x_s;\mathcal{C}), & \text{if attempt $s$ is evaluated},\\
        0, & \text{otherwise}.
    \end{cases}
    \label{eq:scheduled_yield}
\end{equation}
This scheduled-attempt yield combines generation completion and conditional
candidate quality: an attempt contributes only when it reaches evaluation and
satisfies all task constraints. It is the primary end-to-end measure for
comparisons under a fixed attempt budget. Construction and pre-evaluation
failures remain misses for
$\mathrm{GR}$ and $\mathrm{Yield}_{\mathrm{sched}}$, but do not enter the
denominator of $\mathrm{HR}_{\mathrm{eval}}$ or
$\mathrm{Stab}_{\mathrm{eval}}$.

For trajectory analysis, within-run valid-formula redundancy is the fraction
of valid occurrences beyond the first occurrence of each normalized formula
within the same run. It measures exact-formula repetition rather than
crystallographic diversity. Cross-run formula overlap is analyzed separately
in Appendix~\ref{app:formula_repetition}. We report descriptive means and
sample standard deviations where repeated-run aggregates are available and do
not pool the heterogeneous task suite for null-hypothesis testing.

\section{Experimental Instantiation of TRACE}
\label{app:reproducibility}

\paragraph{Search configuration.}
\label{app:experimental_configuration}
Each run begins with ten evaluated initial candidates assigned to five islands
using the role-aware procedure in Appendix~\ref{app:search_state_memory}.
Island selection is balanced for the first 15 iterations and then follows the
adaptive scheduler. Candidate memory is pruned when an island exceeds 5,000
items, retaining higher-scoring successes and low-scoring failures as
contrasting examples for subsequent proposals. For local search, TRACE
enumerates at most 30 candidate edits for a selected parent and forms a
shortlist of five. Learned guidance is activated from iteration 20 only after
at least 20 task-matched evaluated transitions with property deltas have
accumulated. Exact-edit estimates require two observations and are refreshed
every ten iterations. Shortlisted edits use an exploration probability of 0.10
and a softmax temperature of 0.25. Fresh-valid frontier entries remain eligible
for up to 12 iterations, and selected parents enter an eight-iteration cooldown.
Unless stated otherwise, each search iteration contributes two scheduled
candidate-attempt slots, divided equally between global generation and local
editing.

\paragraph{Prompt construction with transition feedback.}
TRACE constructs the context for global proposal generation from candidate and
transition evidence accumulated during search. It includes formulas and
measured properties from the selected island, together with task-level
summaries of executed edits, including support counts, validity rates, residual
gains, and mean property changes. When the transition estimator is active and a
local parent has been scored, TRACE also reports the parent's residual and the
top-ranked edits with their predicted validity and residual progress. These
quantities are used as soft statistical evidence rather than instructions that
the LLM must execute. The formula-generation prompt combines this evidence with
the task objective, property constraints, sampled chemistry rules, and a JSON
schema for candidate formulas and rationales. This prompt defines the global
proposal route, whereas executable local edits remain separately enumerated and
selected by the local route. Each proposed formula is then passed to a
structure-generation prompt that enforces exact stoichiometry and returns the
formula, lattice vectors, atomic species, and fractional coordinates. Both LLM
stages use a temperature of 0.8 and top-$p$ of 1.0.

\begin{table}[H]
    \centering
    \hspace*{-1cm} 
    \footnotesize 
    \setlength{\tabcolsep}{6pt} 
    \renewcommand{\arraystretch}{1.06}
    \caption{Search-efficiency decomposition after 1,000 scheduled attempts.
    Values are percentages summarized over five long-horizon runs per
    task--method pair; metrics follow Appendix~\ref{app:evaluation_metrics}.}
    \label{tab:appendix_efficiency_decomposition}
    \begin{tabular}{@{}l *{6}{c} @{}} 
        \toprule
        & \multicolumn{2}{c}{G.R. $\uparrow$}
        & \multicolumn{2}{c}{Conditional H.R. $\uparrow$}
        & \multicolumn{2}{c}{Scheduled yield $\uparrow$} \\
        \cmidrule(lr){2-3}\cmidrule(lr){4-5}\cmidrule(l){6-7}
        Task & LLEMA & TRACE & LLEMA & TRACE & LLEMA & TRACE \\
        \midrule
        Wide-Bandgap Semiconductors & 80.5 & \textbf{98.2} & 22.6 & \textbf{45.7} & 18.2 & \textbf{44.9} \\
        High-k Dielectrics & 78.2 & \textbf{99.4} & 7.8 & \textbf{29.4} & 6.1 & \textbf{29.2} \\
        Hard, Stiff Ceramics & 82.2 & \textbf{95.1} & 11.8 & \textbf{46.2} & 9.7 & \textbf{43.9} \\
        Perovskite Oxides & 77.3 & \textbf{97.0} & 8.4 & \textbf{21.9} & 6.5 & \textbf{21.2} \\
        \bottomrule
    \end{tabular}
\end{table}
\section{Additional Analyses of Search and Transition Guidance}
\label{app:extended_evidence}

The main experiments establish the end-to-end performance and component-level
contributions of TRACE. This section examines the search dynamics behind these
gains, asking how scheduled attempts are converted into valid discoveries and
how accumulated transition evidence supports local guidance.

\subsection{Sources of Search-Efficiency Gains}
\label{app:search_efficiency_sources}

The cumulative discovery curves in the main paper combine evaluation
completion and constraint satisfaction. A scheduled attempt contributes a
valid discovery only when it reaches property evaluation and the evaluated
candidate satisfies the task constraints. Table~\ref{tab:appendix_efficiency_decomposition}
therefore decomposes the five-run 1,000-attempt endpoints into generation
rate, evaluation-conditional hit rate, and scheduled-attempt yield.

TRACE improved both generation rate and conditional H.R. for every task in the
five-run endpoint summaries. Generation rate increased by $1.16$--$1.27\times$, while
conditional H.R. increased by $2.02$--$3.92\times$, together raising
scheduled-attempt yield by $2.47$--$4.79\times$. Together, these endpoint
values show that the long-horizon discovery gains reflect both higher
evaluation completion and stronger constraint satisfaction among candidates
that reached evaluation.
This endpoint decomposition follows the same five-run long-horizon setting as
the main search-efficiency figure and localizes the mechanism behind the
observed discovery gap. The matched ablations in the main paper isolate
component contributions, and the next subsection examines how accumulated
transition evidence behaves during search.

\subsection{Transition Guidance with Accumulated Evidence}
\label{app:transition_learning_evidence}

We examine accumulated transition evidence from two complementary views. We
first group edits by prior observations of the same exact edit signature, then
follow aggregate outcomes over scheduled-attempt checkpoints. These views
separate edit-level evidence accumulation from the broader temporal evolution
of the search trajectory. Among 882 post-warm-up transitions, 170 had no prior
exact-edit observation, 179 had one, and 533 had at least two, the minimum
support used for an exact-edit estimate. Across the corresponding nonzero
property-change comparisons, direction accuracy increased from 57.0\% to
65.3\% and 75.0\%. Reaching the minimum exact-edit support was therefore
associated with an 18.0-point improvement over edits with no prior exact-match
evidence. This stepwise pattern directly tracks repeated observations of the
same edit signature, while the component ablations in the main paper evaluate
transition history and residual-aware selection at the system level.

Table~\ref{tab:appendix_checkpoint_edits} reports cumulative diagnostics for
post-warm-up edits with observed properties. From 250 to 1,000 attempts, valid
rate increased by 8.14--31.29 points across the four trajectories. Direction
accuracy increased overall on Hard, Stiff Ceramics and Toxic-Free Perovskite
Oxides and varied non-monotonically on the other two tasks, reflecting
task-specific changes in the explored parents and chemical regions. High-k
Dielectrics provides a focused temporal contrast: TRACE trailed LLEMA by 1.97
H.R. points under the repeated short-run protocol, while its direction accuracy
reached 73.55\% for edits with at least two prior exact-edit observations. In
separate long-horizon trajectories, the TRACE--LLEMA gap in cumulative valid
discoveries widened from 8 candidates at 250 attempts to 231 at 1,000 attempts.
This contrast motivates the task-level analysis below, where ranking signal is
examined together with the availability of valid edited children.

\begin{table}[!ht]
    \centering
    \footnotesize
    \setlength{\tabcolsep}{4.8pt}
    \renewcommand{\arraystretch}{1.08}
    \caption{Cumulative edit diagnostics at the scheduled-attempt checkpoints
    used in the main search-efficiency analysis. Values are percentages;
    direction accuracy is computed over nonzero property changes.}
    \label{tab:appendix_checkpoint_edits}
    \begin{tabular}{@{}c@{\hspace{10pt}}*{8}{c}@{}}
        \toprule
        \multirow{2}{*}{\makecell[c]{Scheduled\\attempts}}
        & \multicolumn{2}{c}{\makecell{Wide-Bandgap\\Semiconductors}}
        & \multicolumn{2}{c}{\makecell{High-k\\Dielectrics}}
        & \multicolumn{2}{c}{\makecell{Hard, Stiff\\Ceramics}}
        & \multicolumn{2}{c}{\makecell{Perovskite\\Oxides}} \\
        \cmidrule(lr){2-3}\cmidrule(lr){4-5}
        \cmidrule(lr){6-7}\cmidrule(l){8-9}
        & Dir. acc. & Valid rate
        & Dir. acc. & Valid rate
        & Dir. acc. & Valid rate
        & Dir. acc. & Valid rate \\
        \midrule
        250   & 70.78 & 48.05 & 69.77 & 17.44 & 65.79 & 77.19 & 74.67 & 24.00 \\
        500   & 77.46 & 40.88 & 64.80 & 27.93 & 68.67 & 79.33 & 77.59 & 38.51 \\
        750   & 74.60 & 55.60 & 63.67 & 43.88 & 71.08 & 84.34 & 77.18 & 42.46 \\
        1,000 & 73.17 & 65.53 & 64.47 & 48.73 & 73.20 & 85.33 & 78.61 & 48.84 \\
        \bottomrule
    \end{tabular}
\end{table}

\subsection{Task Dependence and Applicability of Transition Guidance}
\label{app:task_dependence_transition_guidance}

\begin{figure}[H]
    \centering
    \includegraphics[width=\linewidth]{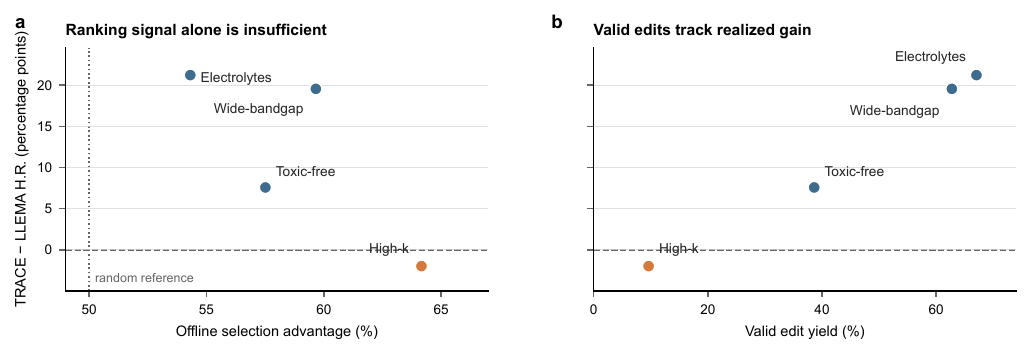}
    \caption{Task-level conditions for realizing gains from transition
    guidance. Each point summarizes five short runs for one task. \textbf{a},
    Offline selection advantage versus matched H.R. difference. \textbf{b},
    Valid-child rate versus matched H.R. difference.}
    \label{fig:task_dependence_transition_guidance}
\end{figure}

We next ask when the transition-ranking signal translates into task-level H.R.
gains. Across the four short-run task aggregates, offline selection advantage
exceeded its 50\% random reference in every case, confirming that transition
history provides useful ranking information. Figure~\ref{fig:task_dependence_transition_guidance}
shows that realized H.R. gains were more closely aligned with the availability
of task-valid edited children. High-k Dielectrics had the largest selection
advantage (64.16\%) but the lowest valid rate (9.62\%) and a $-1.97$-point
H.R. difference, whereas the two tasks with valid rates above 60\% improved
H.R. by approximately 20 points. All quantities are computed from the same five
short runs per task.

Individual transitions show why this task-level pattern arises. Beneficial
edits can close residual constraints, as in
$\mathrm{B_2S_3}\xrightarrow{\mathrm{B}\to\mathrm{Al}}\mathrm{Al_2S_3}$
for Wide-Bandgap Semiconductors, which reduced the normalized residual from
0.298 to zero, and
$\mathrm{CaSrTiO_3}\xrightarrow{\mathrm{Sr}\to\mathrm{Mg}}\mathrm{CaMgTiO_3}$
for High-k Dielectrics, which converted an invalid parent into a valid child.
A related substitution can also move a valid parent away from the feasible
region: replacing Al with B in $\mathrm{MgAl_2O_4}$ produced
$\mathrm{Mg(BO_2)_2}$ and increased the residual from zero to 2.816. Together,
the task aggregates and individual transitions show that aggregate edit effects
act as search priors conditioned by the parent and remaining residuals.
Transition guidance is therefore most useful when the executable edit space
contains reusable changes aligned with the current constraint violations.

\section{Diversity and Search Focus}
\label{app:chemical_space_diversity}

We further examine whether TRACE preserves search diversity while improving
validity, using exact-formula repetition and elemental coverage as complementary
diagnostics.

\subsection{Formula Repetition Across Independent Runs}
\label{app:formula_repetition}

We first compare exact-formula repetition in the matched short-run archives.
Formula strings are normalized by removing formatting characters while
retaining stoichiometry, and within-run repetition is the fraction of evaluated
occurrences beyond the first occurrence of each normalized formula.
Table~\ref{tab:appendix_formula_repetition} reports five-run means and sample
standard deviations for all evaluated and task-valid candidates, together with
the pooled cross-run overlap of valid post-warm-up TRACE edit offspring.

\begin{table}[H]
    \centering
    \footnotesize
    \setlength{\tabcolsep}{5pt}
    \renewcommand{\arraystretch}{1.06}
    \caption{Exact-formula repetition in matched short-run LLEMA and TRACE
    archives. Values are percentages; within-run entries report mean $\pm$
    sample standard deviation over five runs, and the final column reports
    pooled cross-run overlap of valid TRACE edit offspring.}
    \label{tab:appendix_formula_repetition}
    \begin{tabular}{l c c c c c}
        \toprule
        & \multicolumn{2}{c}{All evaluated}
        & \multicolumn{2}{c}{All valid}
        & \multicolumn{1}{c}{} \\          
        \cmidrule(lr){2-3}\cmidrule(lr){4-5}
        Task & LLEMA & TRACE & LLEMA & TRACE & Pooled overlap \\
        \midrule
        Wide-Bandgap Semiconductors
        & $18.74\!\pm\!6.68$ & $1.74\!\pm\!1.13$
        & $24.06\!\pm\!11.31$ & $1.11\!\pm\!1.11$ & 18.25 \\
        High-k Dielectrics
        & $21.21\!\pm\!4.40$ & $4.07\!\pm\!2.09$
        & $10.30\!\pm\!9.43$ & $0.00\!\pm\!0.00$ & 0.00 \\
        Solid-State Electrolytes
        & $10.68\!\pm\!3.49$ & $5.29\!\pm\!2.24$
        & $17.14\!\pm\!6.36$ & $8.00\!\pm\!3.70$ & 10.53 \\
        Perovskite Oxides
        & $24.06\!\pm\!4.22$ & $5.66\!\pm\!3.49$
        & $20.93\!\pm\!9.59$ & $3.41\!\pm\!3.37$ & 14.29 \\
        \bottomrule
    \end{tabular}
\end{table}

TRACE showed lower within-run repetition than LLEMA on every task, both among
all evaluated candidates and among valid candidates. Excluding initial
candidates yielded the same ordering: repetition among non-initial candidates
was 8.13--22.09\% for LLEMA and 1.80--5.18\% for TRACE; among non-initial valid
candidates, the corresponding ranges were 10.30--24.06\% and 0--7.77\%.
Thus, TRACE maintained lower repetition even after removing the shared
initialization formulas used to seed the islands. Because the local route
filters previously evaluated child formulas, cross-run pooled overlap provides
the more informative check for repeated TRACE edit outcomes after the
seen-formula state is reset. Across independently initialized runs, this
overlap ranged from 0 to 18.25\%, indicating that valid edit outcomes remained
distributed across formulas rather than collapsing onto a small shared set.

\begin{figure}[H]
    \centering
    \includegraphics[width=\linewidth]{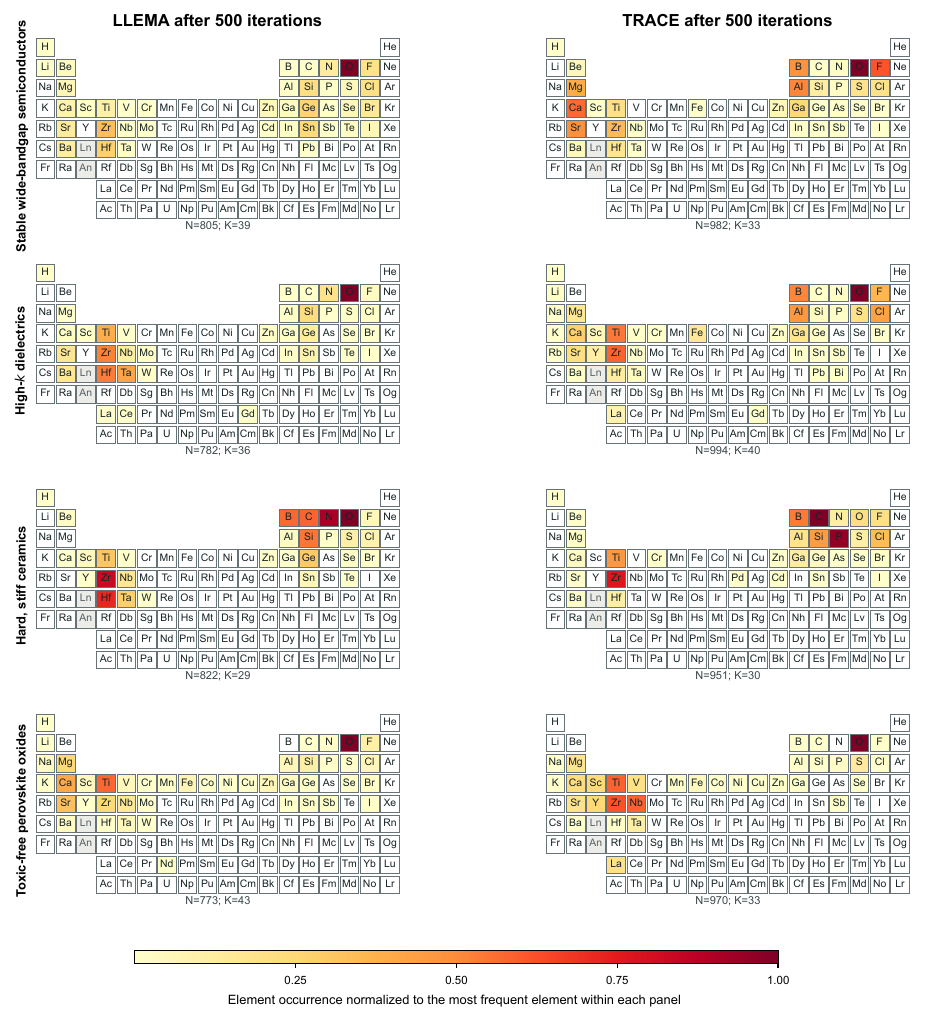}
    \caption{Elemental distributions produced by LLEMA and TRACE at the common
    500-iteration checkpoint. Colors show within-panel relative occurrence;
    $N$ denotes evaluated candidates and $K$ denotes distinct observed
    elements.}
    \label{fig:element_diversity_llema500_trace500}
\end{figure}

\subsection{Elemental Coverage and Task-Adaptive Search Focus}
\label{app:trace_elemental_coverage}

We compare elemental distributions from matched LLEMA and TRACE long-horizon
trajectories at the common 500-iteration checkpoint across the four shared
tasks.
Element occurrences are computed from all evaluated candidates within the
common iteration budget and normalized separately for each task--method pair.

TRACE changed elemental coverage in a task-dependent manner. Relative to LLEMA,
the number of observed elements decreased from 39 to 33 for Wide-Bandgap
Semiconductors and from 43 to 33 for Toxic-Free Perovskite Oxides, increased
from 36 to 40 for High-k Dielectrics, and remained nearly unchanged for Hard,
Stiff Ceramics (29 versus 30). TRACE achieved higher conditional H.R. under
both coverage patterns (Table~\ref{tab:appendix_efficiency_decomposition}).
These patterns support task-adaptive search focus rather than a uniform
tradeoff in which higher validity is obtained by restricting exploration to
fewer elements. Because $K$ measures elemental presence, this analysis
complements the exact-formula repetition results by showing how TRACE shifts
search focus across elements.

\section{Discussion and Limitations}
\label{app:discussion_limitations}

The results suggest that transition history provides a useful intermediate representation between candidate-level feedback and full dynamics modeling. By linking executed edits to observed property changes, TRACE can reuse local search experience without learning a full parent-conditioned transition model. The ablations and transition-level analyses show that these recorded edit-property associations provide predictive and actionable feedback for subsequent search, while global LLM proposals remain important for maintaining broader exploration.

TRACE also has several limitations. Its current estimator captures empirical edit-property regularities across observed transitions, so its effectiveness may decrease when edit effects vary strongly across candidate states; incorporating parent-specific context and predictive uncertainty could improve robustness. The present local edit space is mainly limited to chemistry-constrained, structure-preserving elemental substitutions, and broader operators such as vacancies, lattice changes, or larger structural modifications remain future work. 

\end{document}